\documentclass{article}
\usepackage{spconf,amsmath,graphicx}
\usepackage{amsmath}
\usepackage{amssymb}
\usepackage{amsthm}
\usepackage{enumitem}
\usepackage{enumerate}
\usepackage{balance}
\usepackage{bm}
\usepackage{bbold}
\usepackage{balance}

\usepackage{algpseudocode}
\usepackage{algorithm}     
\usepackage{epstopdf}
\usepackage{subeqnarray}
\usepackage{cases}
\usepackage{color}
\usepackage{cite}
\usepackage{xcolor}

\newtheorem{theorem}{Theorem}

\usepackage{algorithm,algpseudocode}
\algnewcommand{\Initialize}[1]{%
	\State \textbf{Initialize:}
	\Statex \hspace*{\algorithmicindent}\parbox[t]{.8\linewidth}{\raggedright #1}
}
\algnewcommand{\Inputs}[1]{%
	\State \textbf{Inputs:}
	\Statex \hspace*{\algorithmicindent}\parbox[t]{.8\linewidth}{\raggedright #1}
}
\algnewcommand{\Outputs}[1]{%
	\State \textbf{Outputs:}
	\Statex \hspace*{\algorithmicindent}\parbox[t]{.8\linewidth}{\raggedright #1}
}

\title{Federated learning with class imbalance reduction\\
}
\name{\parbox{\linewidth}{\centering
		Miao Yang$^{\dagger\ast}$ \quad Akitanoshou Wong$^{\dagger}$ \quad Hongbin Zhu$^{\dagger}$ \quad Haifeng Wang$^{\ddagger}$ \quad Hua Qian$^{\ast \dagger}$
	}\thanks{This work was supported in part by the National Natural Science Foundation of China (Grant No. 61671436) and the Science and Technology Commission Foundation of Shanghai (Grant No. 19DZ1204300).}}
\address{$^\dagger$School of Information Science and Technology, ShanghaiTech University\\
	$^\ast$Shanghai Advanced Research Institute, Chinese Academy of Sciences (CAS)\\
	$^\ddagger$Key Laboratory of Wireless Sensor Network Communication, CAS\\
	Emails:~\{yangmiao, zhuhb1, qianhua\}@shanghaitech.edu.cn,\\ akidasho.water6@gmail.com, haifeng.wang@wico.sh
}
\begin{document}
	\ninept
	\maketitle
	\begin{abstract}
		Federated learning (FL) is a promising technique that enables a large amount of edge computing devices to collaboratively train a global learning model. 
		Due to privacy concerns, the raw data on devices could not be available for centralized server.
		Constrained by the spectrum limitation and computation capacity, only a subset of devices can be engaged to train and transmit the trained model to centralized server for aggregation.
		Since the local data distribution varies among all devices, class imbalance problem arises along with the unfavorable client selection, resulting in a slow converge rate of the global model.
		In this paper, an estimation scheme is designed to reveal the class distribution without the awareness of raw data.
		Based on the scheme, a device selection algorithm towards minimal class imbalance is proposed, thus can improve the convergence performance of the global model.
		Simulation results demonstrate the effectiveness of the proposed algorithm.
	\end{abstract}
	
	
	\section{Introduction}
	\label{sec:intro}
	With the growing amount of applications,
	end devices, e.g., smart phones, tablets, or vehicles, generate massive private data in daily life \cite{MChiang2016}.
	The valuable personal data can be harnessed to train the machine learning model and significantly improve the quality of end-users' experience.
	However, due to privacy concerns, transmitting private data from local client devices to the cloud server is not feasible and appropriate.
	
	To efficiently utilize the end-users' data, federated learning (FL) has emerged as a new paradigm of distributed machine learning that executes model training with the private data in local devices \cite{konevcny2016federated}.
	Engaging FL, locally training model can be transmitted to global server for model aggregation without any data information.
	Unlike the server-based applications, FL on end devices poses several fundamental challenges, such as the limited connectivity of wireless networks, unstable availability of end devices, and the non identically and independently distributed (non-IID) distributions of client dataset.
	The concerns mentioned above prohibits the model training on all participating devices from beginning to end.
	To avoid such a dilemma, only a subset of devices is selected to participate in each round of model training which is a common practice in FL\cite{lin2017deep}.
	
	In recent years, a variety of client selection schemes in FL have been advocated.
	In \cite{shi2020device}, the authors proposed a client scheduling approach to achieve a proper trade-off between the learning efficiency and latency per round.
	The staleness of the received models and instantaneous channel qualities were jointly considered in \cite{yang2020age}.
	Besides, the work in \cite{zeng2020energy} investigated the problem of minimizing energy consumption of edge devices in FL without compromising learning performance. 
	The client data's distribution of the above FL methods is IID , which may not be applicable in practice.
	
	The device usage pattern of different users varies, so the data samples and labels on any individual device may follow a different distribution.
	A individual data distribution cannot represent the global data distribution.
	Recently, it has been pointed out that the performance of FL,
	especially federated averaging (FedAvg) algorithm \cite{mcmahan2017communication}, may significantly degrade in the presence of non-IID data, in terms of the model accuracy and the communication rounds required for convergence \cite{zhao2018federated,sattler2019robust,li2019convergence}.
	Recognizing such criticality, the authors in \cite{wang2020towards} designed a scheme to mitigate the impact of the class imbalance and introduce a loss function to evaluate the class imbalance.
	Moreover, the work in \cite{wang2020optimizing} exploited learning algorithm for client selection to decrease the communication rounds with target accuracy.
	However, the algorithm in \cite{wang2020optimizing} needs lots of offline training and the data on each device must remain unchanged during the training procedure.
	
	In this paper, the client selection problem concentrated on class imbalance in FL is investigated.
	With the concern of the users' privacy, we propose a scheme that can reveal the severity of class imbalance without any raw data of client devices.
	Besides, utilizing reinforcement learning, we propose a client selection scheme to minimize the effect of class imbalance.
	The proposed algorithm endeavors to learn the class distribution and selects the most balanced clients combination.
	
	
	The remainder of this paper is organized as follows.
	The system model is presented in section \ref{sec:system model}.
	In section \ref{sec:algorithm}, we introduce a scheme to evaluate the class imbalance of clients and propose a learning algorithm to find the best client set.
	The performance of the proposed algorithm is numerically evaluated in section \ref{sec:numerical results}.
	In section \ref{sec:conclusions}, we conclude this paper.

	\section{System Model}
	\label{sec:system model}
	
\begin{figure}
		\includegraphics[scale=0.58]{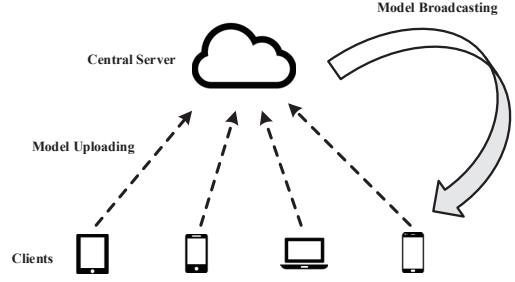}
		\centering
		\caption{An illustration of FL procedure.}
		\label{fig:system_model}
	\end{figure}

	Consider training a deep neural network (DNN) under FL settings with a set of clients $\mathcal{K}=\{1,2,...,K\}$, each with its own local dataset and a global server for model aggregation, as illustrated in Fig.~\ref{fig:system_model}.
	We model the channel from the device to global server as multiple access channel.	
	With the communication limitation, only a fixed amount of spectrum is available \cite{xia2020multi}.
	Due to the scarce of the spectrum resource, the number of available channels is much smaller than that of the client devices.
	
	We then formally introduce the DNN training for multiclass classification problem in FL. 
	Consider a class classification problem defined over a compact feature space $\mathcal{X}$ and a label space $\mathcal{Y}=\mathcal{C}$ with $C$ classes, where $\mathcal{C}=\{1,\cdots,C\}$.
	Let $(\bm{x},y)$ denote a particular labeled sample.
	A function $f:\mathcal{X}\rightarrow\mathcal{D}$ maps $\bm{x}$ to the probability for the $i$th class, where $\mathcal{D}=\{\bm{z}|\sum_{i=1}^{C}z_i=1,z_i\geq 0,\forall i\in\mathcal{C}\}$.	
	Let $\bm{W}$ denote the weight of DNN.
	The cross entropy loss can be harnessed to evaluate the training performance in classification, which is defined as
	\begin{align}
		L(\bm{w}) &=E_{\bm{x},y\sim p}\Big[\sum_{i=1}^{C}\mathbb{1}_{y=i}\log f_i(\bm{x},\bm{W})\Big]\nonumber\\ &=\sum_{i=1}^{C}p(y=i)E_{\bm{x}|y=i}[\log f_i(\bm{x},\bm{W})].
	\end{align}
	where $E(\cdot)$ denotes the expectation operation and $\mathbb{1}$ denotes the indicator matrix.

	In FL, the training procedure is an iterative process consisting of a number of communication rounds.
	Let $\bm{W}^k_t$ and $\bm{W}^g_t$ denote the weights of $k$th device's DNN model and global DNN model in round $t$, respectively.
	In round $t$, the server chooses a subset $\mathcal{S}_t\subseteq \mathcal{K}$ of the clients and then distributes the weights $\bm{W}_{t-1}^g$ of the global model to the selected clients.
	Then the selected clients synchronize their local models such that $\bm{W}_{t-1}^{k}=\bm{W}_{t-1}^g$, and perform the following stochastic gradient descent (SGD) training as
	\begin{align}
	\bm{W}_t^{k}=\bm{W}_{t-1}^{k}-\eta_{t}^{k}\nabla_{\bm{W}}L(\bm{W}_{t-1}^{k}; \xi^{k}),
	\end{align}
	where $\eta_t^{k}$ is the learning rate setting for device $k$ in round $t$, $\xi^{k}$ is an example sampled from local dataset in device $k$. 
	
	Once $\bm{W}_t^{k}$ is obtained, each participating device $k$ updates its own model weight difference $\Delta_t^k$ to the global server, which is defined as
	\begin{align}
	\Delta_t^{k}=\bm{W}_t^{k}-\bm{W}_{t-1}^{k}.
	\end{align}	
	When the global server collects all the updates from client devices, it performs the FedAvg algorithm to update the global model as follows \cite{mcmahan2017communication}
	\begin{align}
	\Delta_t^g&=\sum_{k \in \mathcal{S}_t}\frac{n_k}{\sum_{k'=1}^Kn_{k'}}\Delta_t^{k},\\
	\bm{W}_t^g&\leftarrow\bm{W}_{t-1}^g+\Delta_t^g.
	\end{align}
	
	When the data and label distribution on different devices are IID, FedAvg has been shown to perform well approximating the model trained on centrally data \cite{mcmahan2017communication}.
	However, in practice, data owned by each device are typically non-IID, i.e., the data distributions of clients vary from different devices due to different user preferences and usage patterns.
	When data distributions are non-IID, FedAvg algorithm is unstable and may even diverge \cite{zhao2018federated}.
	
	The problem arises by the inconsistency between the locally performed SGD algorithm, which aims to minimize the loss value on local samples on each device and the global objective of minimizing the overall loss on server data samples.
	Since the training model is fitted on different devices to heterogeneous local data, the divergence among the weights $\bm{W}^{k}$ of these local models will be accumulated and eventually degrades the performance of the learning process \cite{mohri2019agnostic}.

	\section{Online Learning Algorithm for Client selection}
	\label{sec:algorithm} 
	In this section, we first discuss the approach to estimate the class imbalance of each training client.
	Then we propose an online learning algorithm to find the most balanced clients set.
	\subsection{Class Estimation Scheme}
	\label{sec:banalce_detect}
	
	In FL settings, the raw data of clients could not be obtained due to privacy concerns.
	However, with the help of the following scheme, we can reveal the class distribution of client services according to their updated gradients. 
	
%
	
	
	Consider a DNN contains an input layer, a hidden layer and an output layer.
	Note that in DNN for the classification problem, the scale of the output layer is equal to the size of class label.
	That is, each neuron of the output layer corresponding to a class label.
	Let $\bm{\mathcal{W}}=[\bm{w}_1,\bm{w}_2,...,\bm{w}_C]$ denotes the weights between hidden layer and output layer. Every element of $\bm{\mathcal{W}}$ denote the weights connect hidden layer to the specific neuron of output layer.
	
	Without loss of generality, we assume there exists a balanced auxiliary dataset with $C$ classes in global server. 
	Such auxiliary dataset only consists of a few data samples and can be extracted from the test dataset.
	When the auxiliary data examples are fed to the updated model, we can obtain the gradients vector brought by auxiliary data with respect to the corresponding classes. The gradients vector brought by auxiliary data can be expressed as 
	\begin{align}
		\Big\{\nabla L^{\textit{aux}}(\bm{w}_1), \nabla L^{\textit{aux}}(\bm{w}_2), ... , \nabla L^{\textit{aux}}(\bm{w}_C)\Big\}\nonumber,
	\end{align}	
	where each $\nabla L^{\textit{aux}}(\bm{w}_i)$ is related to the $i$th neuron and class $\mathcal{C}_i$.
	The following theorem can help to obtain the class distribution from the above gradients vector.
	\begin{theorem}
		\label{theorem2}
		When training DNN in classification problem, the expectations of gradient square for different classes have the following approximate relation \cite{anand1993improved}:
		\begin{align}
		\frac{E||\nabla L(\bm{w}_i)||^2}{E||\nabla L(\bm{w_}j)||^2}\approx \frac{n_i^2}{n_j^2},
		\end{align}
		where $L$ denotes the cost function of the neural network, $n_i$ and $n_j$ are the number of samples for class $i$ and class $j$ , respectively, where $i \neq j$ and $i, j \in \mathcal{C}$.
	\end{theorem}
	\textbf{Theorem 1} reveals the correlation between the gradients and class distribution.
	Then for class $\mathcal{C}_i$, the estimation of class ratio $\frac{n_i^2}{\sum_j n_j^2}$ can be defined	
	\begin{align}
		R_{i} = \frac{e^{\frac{\beta}{||\nabla L^{\textit{aux}}(\bm{w}_i)||^2}}}{\sum_{j}e^{\frac{\beta}{||\nabla L^{\textit{aux}}(\bm{w}_j)||^2}}},
		\label{equ:ratio}
	\end{align}	
	where $\beta$ is a hyperparameter that can be tuned to control the normalization between classes. 
	Then we can obtain the composition vector $\bm{R}=[R_1,...,R_C]$ that indicates the distribution of raw data.
	Moreover, the Kullback-Leibler (KL) divergence can be harnessed to evaluate the class imbalance of each client, which is defined as
	\begin{align}
	D_{KL}(\bm{R}||\bm{U}) = \sum_{i\in\mathcal{C}}R_i\log\frac{R_i}{U_i},
	\end{align}
	where $\bm{U}$ is a vector of ones with magnitude $C$.

	\subsection{Online Learning Framework for Client Selection}
	Multi-arm bandit problems are motivated by a variety of real-world problems, such as online advertising, dynamic pricing and stock investment.
	To find the optimal balance client subset, the difficulty lies in how to learn the statistics of class distribution via iterative updated gradient.
	Combinatorial multi-Armed bandit (CMAB) can be harnessed to solve this problem \cite{chen2013combinatorial}.
	In CMAB model, an agent gambles on a bandit machine with a finite set of arms, where each arm has unknown distribution.
	At round $t$, several arms defined as a super arm $\mathcal{S}_t$ ($\mathcal{S}_t\subseteq \mathcal{K}$) can be pulled.
	The reward of the super arm depends on the outcomes of all pulled arms.
	
	In this work, we consider the client selection as a CMAB problem, where each client represents the arm and the client set represents the super arm.	
	In FL training, once the model is updated, the server could obtain the local model of each client device.
	Utilizing the class estimation scheme, we can reveal the composition vector $\bm{R}^k$ of selected client $k$.
	Define the reward of client $k$ as
	\begin{align}
		r^k = \frac{1}{D_{KL}(\bm{R}^k||\bm{U})}\label{equ:reward}.
	\end{align}
	Then the global server executes FedAvg algorithm for model aggregation and the reward $r$ of the whole client set can be obtained the same as (\ref{equ:reward}).
	Note that the super arm's reward is a nonlinear combination of the selected single arms' reward.
	An algorithm that has good theoretical results with nonlinear reward is the combinatorial upper confidence bounds (CUCB) algorithm \cite{chen2013combinatorial}.
	
	Let $T^k$ denote the number of times that the client $k$ has been selected.
	Once client $k$ has been selected in a time slot, $T^k\rightarrow T^k+1$, otherwise, $T^k\rightarrow T^k$.
	The proposed client selection algorithm based CUCB is shown in \textbf{Algorithm 1}.
	\begin{algorithm}[h]
		\caption{CUCB for Client Selection}
		\label{alg:CUCB}
		\begin{algorithmic}[1]
			\State For each client $k$, choose an arbitrary set $\mathcal{S}\in\mathcal{K}$ such that $k\in\mathcal{S}$ and update variables $T^k$ and $\hat{r}^k$.
			\State $t\leftarrow N$.
			\While{\textbf{true}}
			\State $t\leftarrow t+1$.
			\State For each client $i$, set $\hat{r}^k=\bar{r}^k+\alpha\sqrt{\frac{3\ln t}{2T^k}}$.
			\State Obtain $\mathcal{S}_t$ using \textbf{Algorithm 2} with $\hat{r}$.
			\State Play $\mathcal{S}_t$ and update $T^k$ and $\hat{r}^k$.
			\EndWhile
		\end{algorithmic}
	\end{algorithm}

	Here $\alpha$ is the exploration factor to balance the trade-off between exploitation and exploration.
	Step $1$ of \textbf{Algorithm 1} guarantees that each client has been selected once at least in the first $N$ round.
	Notation $\bar{r}^k$ denotes the individual reward sample mean of client $k$, and $\hat{r}^k$ denotes the perturbed version of $\bar{r}^k$.
	In step $6$, the proposed algorithm utilizes the perturbed $\hat{r}^k$ to solve the client selection problem.
	The perturbation in step $5$ promotes the selection of clients that are not selected frequently, by artificially increasing their expected reward estimates. 

	In our client selection problem, the class distribution of each client is uncertain.
	Fortunately, we can reveal the class distribution according to the composition vector.
	Let $\bm{R}^k(t)$ represent the composition vector of client $k$ at time slot $t$.
	Thus, the class ratio can be estimated by the sample mean of composition vector, which can be expressed as
	\begin{align}
		\bar{\bm{R}^k} = \frac{\sum_{t=1}^{T^k}\rho^{T^k-t}\bm{R}^k(t)}{\sum_{t=1}^{T^k}\rho^{T^k-t}},
	\end{align}
	where $\rho$ is the forgetting factor since the characteristic of client class distribution may vary at each time slot.
	
	With the estimated composition vector $\bar{\bm{R}}$ and reward $r$ of each client, we can design the client selection scheme with minimal class imbalance according to \textbf{Algorithm 2}.
	\begin{algorithm}[]
		\caption{Class Balancing Algorithm}
		\label{alg:CS}
		\begin{algorithmic}[1]
			\Initialize{Set $\mathcal{S}_t=\emptyset$ and $\bm{R}_{total}=\emptyset$.
			}
			\State $k_0=\arg\max_k \hat{r}^k$.
			\State $\mathcal{S}_t\leftarrow \mathcal{S}_t\cup\{k_0\}$.
			\While{$|\mathcal{S}_t|<K$}
			\State Select $k_{min}=\arg\min_{k}D_{KL}\Big((\bm{R}_{total}+\bar{\bm{R}}^k)||\bm{U}\Big)$ for $k\in\mathcal{K}\setminus \mathcal{S}_t$.
			\State Set $\mathcal{S}_t\leftarrow \mathcal{S}_t\cup\{k_{min}\}$, $\bm{R}_{total}\leftarrow \bm{R}_{total}+\bar{\bm{R}}^k_{min}$.
			\EndWhile
			\State \textbf{Outputs: }$\mathcal{S}_t$
		\end{algorithmic}
	\end{algorithm}

	\textbf{Algorithm 2} determines the participants of FL updating.
	Combine \textbf{Algorithm 1} and \textbf{Algorithm 2}, we can find the most suitable client set with class balance.
	
	\section{Numerical Results}
	\label{sec:numerical results}
	In this section, we present numerical results to validate the effectiveness of the proposed algorithms \footnote{The source code of our work can be found in https://github.com/ym1231/fl-cir.}.
	
	We test our scheme on one of the main benchmarks: CIFAR10 \cite{C19}. 
	CIFAR10 dataset consists of $50000$ training examples and $10000$ testing examples of $32 \times 32$ RGB images, categorized by total $10$ classes. 
	The number of clients in FL training is set as $100$.
	To model the imbalanced class distribution, we split the whole CIFAR10 dataset to each client with random amount of classes and random amount of data samples.
	
	The architecture of our deep model is a standard convolutional neural network (CNN), which comprises $3$ convolutional layers followed by rectified linear units (ReLU) nonlinear activations and max-pooling layer and $2$ fully connected layers, with totally $122570$ parameters. 
	Such standard model can meet our needs to validate the effectiveness of our scheme.
	The training data is preprocessed by standard techniques for data augmentation, such as cropping, flipping, changing the color, etc.
	We use standard SGD as our optimizer.
	The learning rate and the learning rate decay of SGD are set as $0.1$ and $0.996$, respectively.
	In each training round, the selected clients train their local models for $5$ epochs.
	The client selects $10$ batches with batch size $10$ at each training epoch.
	
	In our simulation, we set exploration factor $\alpha$ as $0.2$ and forgetting factor $\rho$ as $0.99$.
	We fix the normalization hyperparameter $\beta$ as $1$.
	To evaluate the performance of the proposed algorithm, we compare the performance of the proposed algorithm to the following schemes. 
	(i) Greedy scheme: global server selects the client set with the sample mean information according to the class estimation method;
	(ii) Random scheme: global server randomly selects the client set.
	In addition, we compare the proposed algorithm in IID setting to show the performance gap.
	Note that in IID setting, the class distribution and the number of data samples in each client are set as the same.
	Thus the above selection schemes are the same in practice.

	\begin{figure}[t]
		\includegraphics[width=72mm]{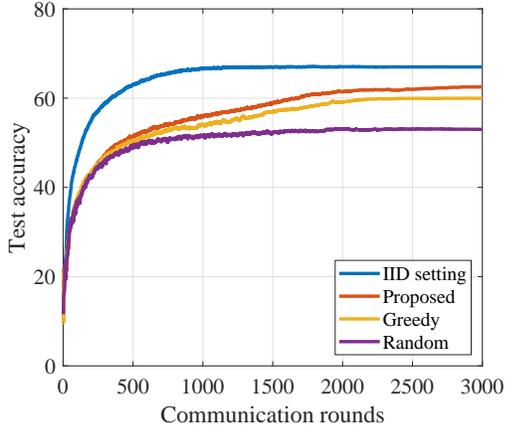}
		\centering
		\caption{The global test accuracy with different selection schemes.}
		\label{fig:diff_algo}
	\end{figure}
	In the first experiment, we examine the convergence performance of global model with the proposed algorithm.
	At each round, the global server selects $20$ clients for model aggregation.
	Fig.~\ref{fig:diff_algo} depicts the test accuracy of global model with different schemes.
	The test accuracy performance represents the learning process of global model.
	From Fig.~\ref{fig:diff_algo}, we observe that the proposed algorithm achieves faster convergence speed and higher test accuracy compared with the greedy and uniform schemes.
	The discrepancy of performance between the proposed and random scheme comes from the effect of class imbalance.
	The proposed algorithm can reduce the class imbalance by carefully selecting the proper client set.
	Compared with the greedy scheme, the proposed algorithm can balance the trade-off between exploration and exploitation, thus resulting in the exploration of more suitable clients set with balanced class combination.

\begin{figure}[]
		\includegraphics[width=72mm]{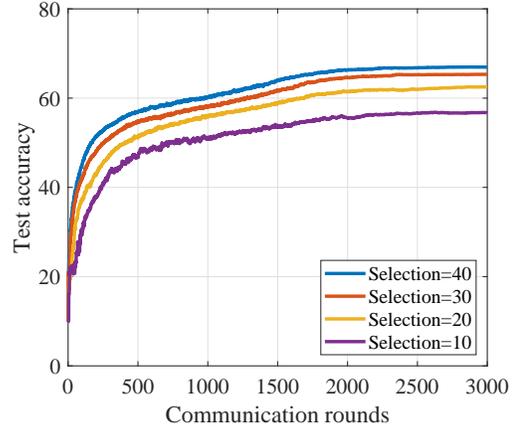}
		\centering
		\caption{The global test accuracy of the proposed algorithm with different amounts of clients selection.}
		\label{fig:diff selection}
	\end{figure}

	In the second experiment, we would like to verify the performance of the proposed algorithm with different amounts of selected clients.
	Fig.~\ref{fig:diff selection} plot the FL training performance with respect to the amounts of clients.
	With the increasing amounts of selected clients, the FL training process can achieve better performance.
	However, the performance improvement is slighter when the amount of clients become larger.
	This result indicates that too large amount of clients is not essential to find the best client set when suffering the burden of communication consumption.

	\begin{figure}[h]
		\includegraphics[width=72mm]{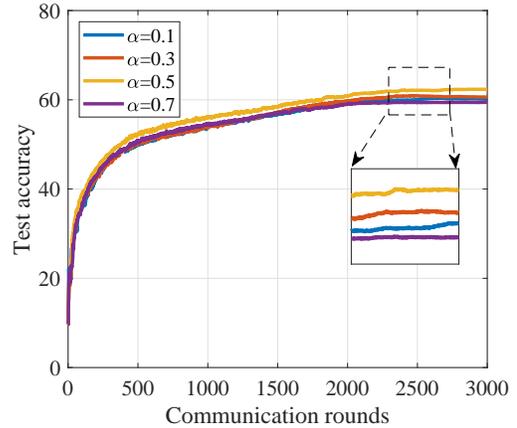}
		\centering
		\caption{The global test accuracy with different exploration factors.}
		\label{fig:diff_alpha}
	\end{figure}
	We investigate the accuracy performance of the proposed algorithm with different parameters $\alpha$ in Fig.~\ref{fig:diff_alpha}.
	The parameters $\alpha$ can decide the trade-off between exploitation and exploration.
	When the exploration parameter $\alpha$ is small, the global server would like to exploit the history sampling information and thus could not explore sufficiently to find the best client set. 
	As the exploration parameter $\alpha$ increases, the global server prefers to explore the clients with fewer selections.
	Spending many rounds for exploration could result in the performance deterioration since the clients selected for exploration may be not always proper.
	A suitable $\alpha$ is essential to improve the convergence performance.

	\section{Conclusions}
	\label{sec:conclusions}
	In this paper, we studied the client selection problem with class imbalance in FL.
	Without the requirement of clients' data information, we designed a scheme to explicitly reveal the class distribution according to the updated gradients.
	Besides, a client selection algorithm based on the CMAB framework was proposed to reduce the class imbalance effect.
	Numerical results confirmed that the proposed algorithm could pick the properly balanced client set and improve the convergence performance of the global model.

	\balance
	\bibliographystyle{IEEEbib}
	\bibliography{strings,refs}
	
\end{document}